\pdfoutput=1

\documentclass[11pt]{article}

\usepackage[]{ACL2023}

\usepackage{times}
\usepackage{latexsym}
\usepackage{array} 
\usepackage{arydshln} 
\usepackage[T1]{fontenc}

\usepackage[utf8]{inputenc}

\usepackage{microtype}
\usepackage{subcaption}

\usepackage{inconsolata}

\usepackage{lipsum}  
\usepackage{amsmath}
\usepackage{mathtools}
\usepackage{amsfonts}

\usepackage{tabularx}
\usepackage{adjustbox}
\usepackage{geometry}
\usepackage{booktabs}
\usepackage{amssymb}
\usepackage{graphicx}
\usepackage{multirow}
\usepackage{pifont}

\usepackage{hyperref}

\title{Empowering Sentence Encoders with Prompting and Label Retrieval for Zero-shot Text Classification}

\newcommand{\ours}{RaLP}

\author{
    Jimin Hong\textsuperscript{\rm * 1,2} 
    Jungsoo Park\textsuperscript{\rm * 1} 
    Daeyoung Kim\textsuperscript{\rm * 1,2} 
    Seongjae Choi\textsuperscript{\rm 1} 
    Bokyung Son\textsuperscript{\rm 1} 
    Jaewook Kang\textsuperscript{\rm 1} \\
    \textsuperscript{\rm 1}NAVER, \textsuperscript{\rm 2}KAIST \\
    \texttt{\{jspark.93, seongjae.choi, bo.son, jaewook.kang\}@navercorp.com}\\
    \texttt{\{jimmyh, daeyoung.k\}@kaist.ac.kr}
}

\begin{document}
\maketitle

\providecommand{\js}[1]{
    {\protect\color{purple!50!orange}{[Jungsoo: #1]}}
}
\providecommand{\sj}[1]{
    {\protect\color{violet}{[Seongjae: #1]}}
}
\providecommand{\jm}[1]{
    {\protect\color{blue}{[Jimin: #1]}}
}
\providecommand{\dy}[1]{
    {\protect\color{teal}{[DY: #1]}}
}
\providecommand{\updated}[1]{
    {\protect\color{red!80!orange}{#1}}
}

\providecommand{\bokyung}[1]{
    {\protect\color{cyan}{[bokyung: #1]}}
}

\providecommand{\marco}[1]{
    {\protect\color{red}{[marco: #1]}}
}

\begin{abstract}
With contrastive pre-training, sentence encoders are generally optimized to locate semantically similar samples closer to each other in their embedding spaces. In this work, we focus on the potential of their embedding spaces to be readily adapted to zero-shot text classification, as semantically distinct samples are already well-separated. Our framework, \ours~(\textbf{R}etrieval \textbf{a}ugmented \textbf{L}abel \textbf{P}rompts for sentence encoder), encodes \emph{prompted} label candidates with a sentence encoder, then assigns the label whose prompt embedding has the highest similarity with the input text embedding. 
In order to compensate for the potentially poorly descriptive labels in their original format, \ours~\emph{retrieves} sentences that are semantically similar to the original label prompt from external corpora and use them as additional pseudo-label prompts.
\ours~achieves competitive or stronger performance than much larger baselines on various closed-set classification and multiple-choice QA datasets under zero-shot settings. We show that the retrieval component plays a pivotal role in \ours's success, and its results are robustly attained regardless of verbalizer variations.

\end{abstract}

\section{Introduction}\label{sec:intro}



\begin{figure}[t]
\centering
\includegraphics[width=0.8\linewidth]{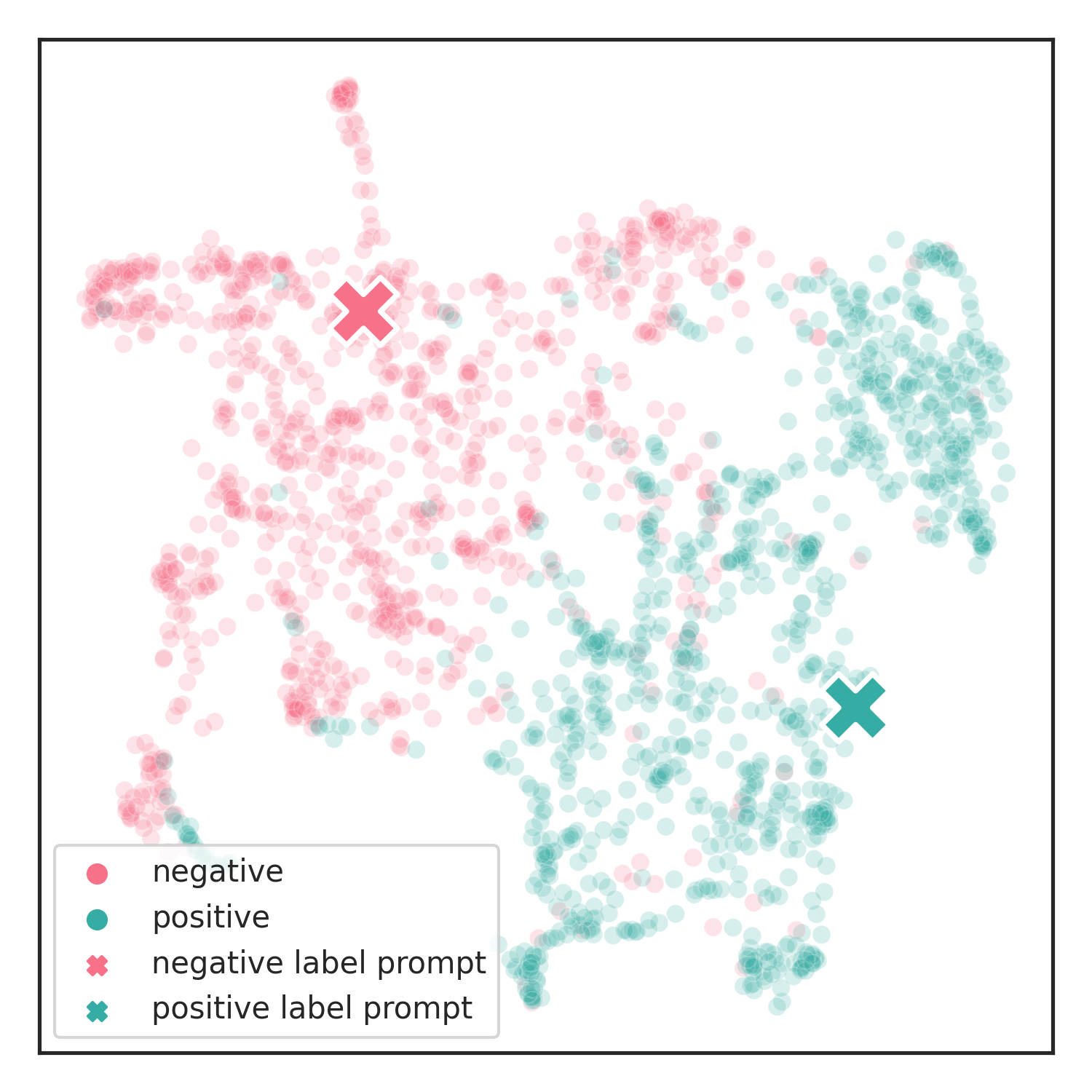}
\caption{
t-SNE~\citep{van2008visualizing} visualization of encoded test samples and label prompts from the CR dataset~\citep{hu2004mining} using a Sentence-t5 encoder~\citep{ni2022sentence}. 
Each colored circle illustrates a sample with its class information ("negative" or "positive"). 
The label prompts for each class (\emph{e.g.} ``It was great.'' and ``It was terrible,'') are shown in bold crosses.
}
\label{fig:intro}
\vspace{-3mm}
\end{figure}

Sentence encoders have been widely applied to a comprehensive range of natural language processing tasks, including classification, semantic retrieval, and semantic similarity tasks~\citep{reimers2019sentence, du2021self, gao2021simcse, ni2022sentence}.
They are usually pre-trained with a contrastive objective on datasets that focus on sentence semantics (\emph{e.g.} NLI), so semantically similar texts are located close to each other in their embedding spaces. 
We note that embedding spaces with such traits could be particularly friendly to classification tasks under limited supervision, as semantically distinct samples are well-separated in advance of any refinement made during downstream training. 
Recent work has demonstrated the competitiveness of rich text embeddings from sentence encoders for few-shot classification~\citep{tunstall2022efficient}, being on par or outperforming much larger prompt-based generative language models on the RAFT benchmark~\citep{alex2021raft}. 
However, it essentially involves training a logistic classification head, which introduces additional parameters to be tuned and is thus inapplicable to zero-shot inference.

In this work, we continue to explore the power of sentence encoders, but this time to solve zero-shot classification by combining with prompting~\citep{brown2020language} and dense retrieval techniques. 
We transform closed or open-set classification tasks into finding the label texts (\emph{i.e.} prompts) with maximal textual similarity with the sentences under inference in a sentence encoder's embedding space. 
Figure~\ref{fig:intro} illustrates a motivational case for our approach, where samples are generally located closer to the label prompt representation of their corresponding class than the other.
Meanwhile, we note that label prompts in their original format are potentially ambiguous or poorly descriptive; for example, \emph{"Topic: World"} in the AGNews dataset is an overly compressive and abstract label, being less adequate to embrace a large range of belonging texts. We handle this issue by using additional support from multiple pseudo-label prompts retrieved from an external corpus.

Prompting has been commonly used for few-shot and zero-shot tasks within the form of in-context learning with generative language models~\citep{brown2020language, gao2021making}. 
However, their performances show high variance across different verbalizers (\emph{i.e.} text expressions for labels), even being close to chance-level~\cite {perez2021true, lu2022fantastically}. 
Unlike generative models, where the label prompts and the query texts are intertwined by concatenation at the input level, sentence encoders isolate the two at encoding time. We empirically show that such post-hoc interactions can be robust to label prompt variations, and the robustness further increases when we enlarge the support prompt set through retrieval.

We conduct extensive experiments on six closed-set classification datasets and six open-set multiple-choice datasets. 
Our best model based on a powerful sentence encoder \emph{Sentence-T5}~\citep{ni2022sentence} with retrieval augmented label prompts achieves comparable or even stronger performance compared to strong zero-shot baselines for the closed-set classification tasks. 
The best results for the multiple choice QA tasks are less impressive; nevertheless, we still reach higher scores than baselines of a similar or somewhat bigger size. 
Our method is consistently strong across verbalizer variations and scales to different model sizes.
Notably, using retrieval-augmented label prompts leads to significant performance gains in closed-set classification, serving as a simple yet promising way to leverage external knowledge resources.

\section{Background}\label{sec:background}We explain the concept of sentence encoders and prompting that we use in this work.

\subsection{Sentence Encoders}

Given a sentence $X_i \in \mathcal{X}$ , a sentence encoder $E$ encodes the sentence into a fixed size embedding vector $\mathbf{h}_i = E(X_i) \in \mathbb{R}^{d}$,
where $\mathcal{X}$ is the set of all natural language texts and $d$ is the preset embedding dimension.
Sentence encoders are commonly trained with the contrastive learning objective~\citep{chen2020simple} with in-batch negatives~\citep{chen2017sampling, henderson2017efficient}.
The loss function pulls the positive pair representation closer to the input representation while pushing away the negatives in embedding space:
\vspace{-4mm}

\begin{equation}
\label{eq:contrastive_loss}
\ell=-\sum_{i=1}^N \log \frac{e^{\operatorname{sim}\left(\mathbf{h}_{i}, \mathbf{h}_i^{+}\right)}}{\sum_{j=1, j\ne i}^N e^{\operatorname{sim}\left(\mathbf{h}_{i}, \mathbf{h}_j\right)}}
\end{equation}

\noindent where $\operatorname{sim}(\cdot)$ is the similarity function, $\mathbf{h}_i^+$ denotes the positive pair for $\mathbf{h}_i$, and $\mathbf{h}_j$ refers to all other instances except $\mathbf{h}_i$ in the batch of size $N$.


\subsection{Prompting}

Given a text input $X_i \in \mathcal{X}$ and a set of labels $\mathcal{Y}=\left\{y_1, \ldots, y_m\right\}$, a predefined verbalizer\footnote{ Consists of templates and label names.} $v: \mathcal{Y} \rightarrow \mathcal{X}$ generates a label prompt in natural language for each label index.
Generally, language models compute the distribution of label prompts given the input, $P_{\mathrm{LM}}\left(v\left(y_m\right) \mid X_i\right)$~\citep{brown2020language}.
Other lines of works utilize channel modeling $P_{\mathrm{LM}}\left(X_i \mid v\left(y_m\right)\right)$~\citep{min2022noisy, min2021metaicl} or masked language modeling objective~\citep{gao2021making}.

\section{Method}\label{sec:method}\begin{figure*}
\centerline{\includegraphics[width=\textwidth]{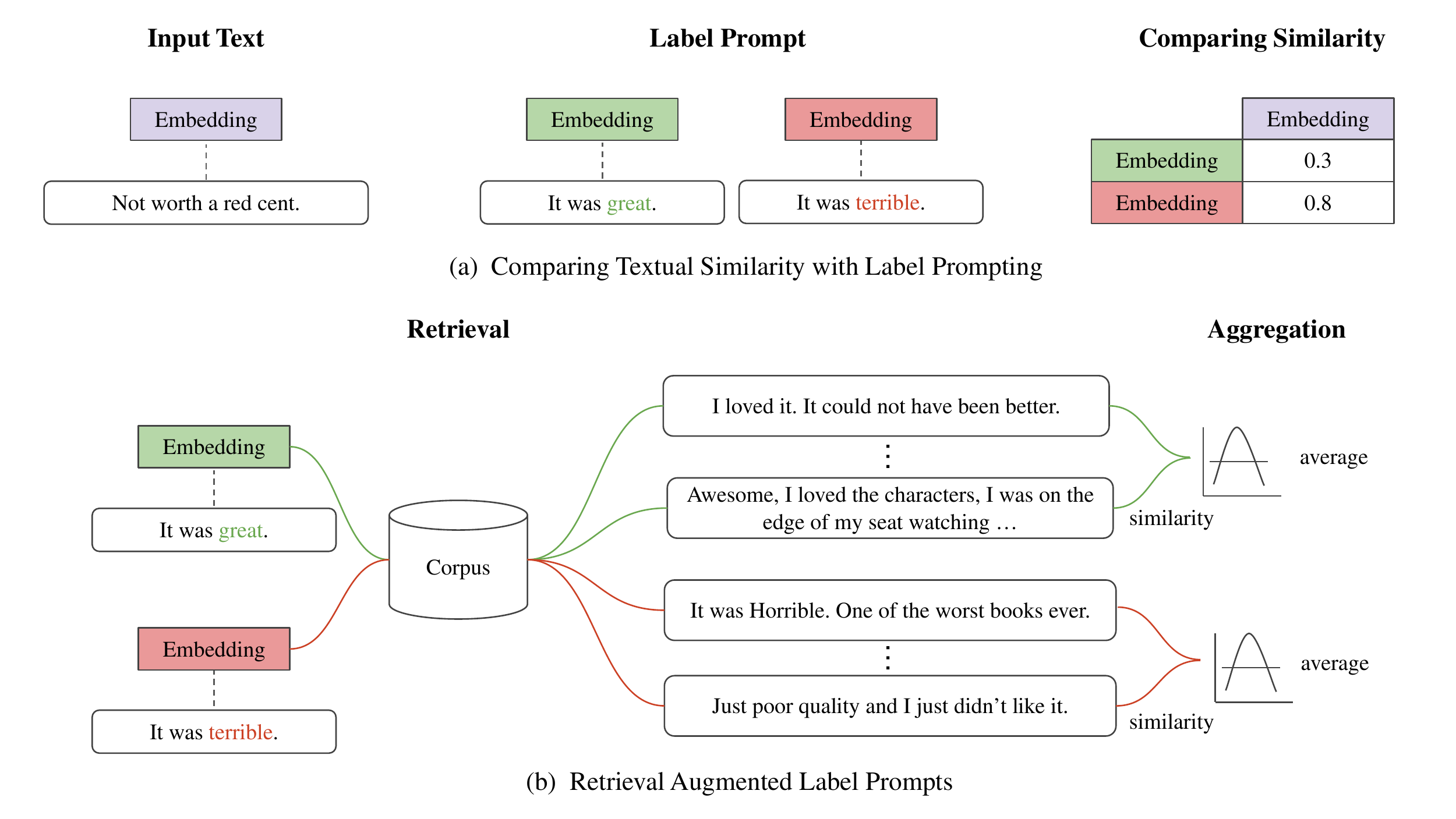}}
\caption{An overview of our approach to a text classification task (or input-label matching). Among a set of label candidates, we find the label whose prompt has the highest representational similarity with the input text in the embedding space of a sentence encoder (Figure (a)). The similarity scores are computed across multiple retrievals augmented label prompts which are collected with original label prompts as queries (Figure (b)).
}
\label{fig:method}
\end{figure*}

We explore solving zero-shot classification tasks using text representations from sentence encoders.
Figure~\ref{fig:method} provides an overview of our approach. 
The gist is to exploit the representational similarity between the input text and retrieval-augmented label prompts.

\subsection{Representational Similarity with Sentence Encoders}
\label{sec:method_tess}
Taking the example in Figure \ref{fig:method}~(a), for a binary sentiment classification task with $\mathcal{Y} = \{y+, y-\}$ where $y+$ stands for "positive" and $y-$ stands for "negative", we first use a verbalizer to change each label into prompts: say, \textit{``It was great.''}$(v(y+))$ and \textit{``It was terrible.''}$(v(y-))$.
Afterward, the label prompts are transformed into encoded vectors that act as prototypes for representing each class cluster: $\mathbf{z}_m = E(v(y_m))$.
Then, we use the following scoring function for each candidate label, defined as its similarity with the text under inference $X_i$ \textit{``Not worth a red cent.''}:

\begin{align}
\vspace{-3mm}
\label{eq:similarity}
& \underset{y_m \in \mathcal{Y}}{\operatorname{argmax}}~~\mathrm{P}_{\mathrm{TS}}(y_m|X_i) \nonumber \\
\mathrm{P}_{\mathrm{TS}}(& y_m| X_i) \propto \operatorname{sim}({\mathbf{h}_i, \mathbf{z}_{m}}) 
\end{align}

\noindent where we use cosine similarity in a sentence encoder's embedding space for $\operatorname{sim}(\cdot)$.


Note that using Eq.~\ref{eq:similarity} at inference time aligns the downstream task objective with the contrastive pre-training objective (Eq.~\ref{eq:contrastive_loss}), which may benefit sentence encoders in a limited label regime.
Plus, separately encoding label prompts from the test input makes the results more robust under verbalizer variations, which we show in Section~\ref{subsection:sensitivity}.

\subsection{Retrieval Augmented Label Prompts}
\label{sec:method_ralp}

It is crucial for the label prompts to reliably express the necessary class information if each of them is to act as the single similarity anchor for its class.
Unfortunately, this precondition often does not hold as many label names are being compressive rather than fully descriptive; \textit{e.g.} \textit{``Topic: World''} in the AGNews dataset. 

To make amends for the potential issue of fragmented semantics in the labels in their original format, we augment the label prompts with semantically similar sentences retrieved from external knowledge sources. 
Texts collected from the wild are expectedly more descriptive. Therefore, by serving as alternative formats or augmentations to the original label prompts, they may add useful class-related information, such as synonyms or related expressions.

We retrieve top-$K$ sentences from an external corpus according to the representational similarity with the given label prompt. Then, we compute the aggregated score for the retrieved \emph{set of sentences} as

\vspace{-4mm}
\begin{align}
\label{eq:external_similarity}
\mathrm{P}_{\mathrm{ES}}(y_m| & X_i) \propto \frac{1}{K}\sum_{k=1}^{K}{\operatorname{sim}({\mathbf{h}_i, \operatorname{topk}(\mathbf{z}_{m}}})^k) 
\end{align}

\noindent where $K$ refers to preset retrieval size, $\operatorname{topk}$ is an operator which returns top-k nearest embeddings from encoded external sources, and $^k$ denotes the k-th element from the retrieval results.
The sum of Eq.~\ref{eq:similarity} and Eq.~\ref{eq:external_similarity} now becomes our new inference scoring function.

\vspace{-4mm}
\begin{equation}
\label{eq:final_sum}
\underset{y_m \in \mathcal{Y}}{\operatorname{argmax}}~~\mathrm{P}_{\mathrm{TS}}(y_m|X_i) + \mathrm{P}_{\mathrm{ES}}(y_m|X_i) \nonumber
\end{equation}

For further augmentation during retrieval, we experiment with using synonyms of the original label names (\textit{e.g.} \textit{``good''}, \textit{``remarkable''} for \textit{``great''}) as in~\citet{shi2022nearest}. Given a label prompt, we first generate a size-$N$ list of synonymous label prompts by replacing the label name with its synonyms. Then, we do the retrieval for $N$ times, each time with a formerly generated synonymous prompt as the query and $K/N$ retrieved sentences. 
\begin{table*}[t]
\centering
\small
\renewcommand{\arraystretch}{1.1}
\begin{tabular}{lrcccccccc}
\toprule
\textbf{Method} & \textbf{\# Params} & \textbf{SST-2} & \textbf{MR} & \textbf{CR} &  \textbf{RT} & \textbf{Yahoo} & \textbf{AGNews} & \textbf{Avg}\\
\hline
\textbf{Baselines (Language Modeling)} \\
GPT-2$^\dagger$ & 774M & 55.3& 54.6        & 66.2        & 53.0        & 49.7& 67.4     &57.7   \\
+ PMI$^\dagger$~\citep{holtzman2021surface} & 774M & 76.5& 74.6        & 82.8        & 74.1        & 48.8& 65.1      &70.3  \\
GPT-3$^\dagger$ & 175B & 63.6 & 57.4        & 53.8        & 57.0        & 53.1 & 75.4      &60.1  \\
+ PMI$^\dagger$~\citep{holtzman2021surface} & 175B & 71.4 & 76.3        & 70.0        & 75.5        & 54.7 & 74.7      &70.4  \\
LM-BFF~\citep{gao2021making} & 355M & 83.2& 80.8        & 79.6        & \textbf{82.6}        & 44.9& 68.3   &73.2   \\

\hline
\textbf{Baselines (Retrieval-Augmented LM)} \\
GPT-2 + kNN$^\dagger$~\citep{khandelwal2019generalization} & 774M & 55.4& 56.4        & 67.2        & 54.5        & 49.5& 67.0    &58.3   \\
GPT-2 + kNN-Prompt$^\dagger$~\citep{shi2022nearest}      & 774M & 84.2& 78.2        & 84.3        & 80.6        & 51.0& \underline{78.8}      &76.2  \\
\hline
\textbf{Baselines (NLI format)} \\
Entailment~\citep{yin2019benchmarking} & 355M & 83.7& 79.6        & 83.8        & 78.2        & 46.0& 75.1  &74.4 \\
\hline
\textbf{Ours (Sentence Encoder)} \\
\ours$_\text{SimCSE}$ & 355M & 82.3 & 78.0        & 87.1        & 76.7        & 57.0 & 72.6    &75.6 \\
\ours$_\text{ST5}$               & 335M & \textbf{87.8} & \textbf{81.7}        & \textbf{87.4}   & 82.4        & \underline{\textbf{57.4}} & \textbf{76.6}   &\textbf{78.9} \\
\ours$_\text{ST5-XL}$               & 1.24B & 88.6 & 82.8 & 86.1 & 83.3 & 56.2 & 75.1 & 78.7 \\
\ours$_\text{ST5-XXL}$              & 4.8B  & \underline{90.5} & \underline{84.7} & \underline{87.7} & \underline{85.1}	& 54.9 & 75.4 &	\underline{79.8} \\
\bottomrule
\end{tabular}
\caption{Results for closed-set classification tasks under the zero-shot setting. 
The \textbf{Boldface} indicates the best performances in each column except the models with over 1B parameters.
The \underline{Underline} shows the best scores in each column regardless of model size.
Values with $\dagger$ are taken from \citet{min2022nonparametric}.
}
\label{tab:main_close_set}
\vspace{-3mm}
\end{table*}
\section{Experimental Settings}\label{sec:experiments}

We experiment with two types of tasks: closed-set classification and multiple choice QA. Both involve choosing the corresponding label given a finite label set for each test sample; what tells them apart is whether all test instances share the same label set (closed-set classification) or each test instance is presented with a different each label set (multiple choice QA).

\subsection{Closed-set Classification Datasets}
\label{subsection:closedset_classification}
\paragraph{Sentiment Analysis} SST-2~\citep{socher2013recursive}, MR~\citep{pang2004sentimental}, CR~\citep{hu2004mining}, and Rotten Tomatoes (RT)~\citep{socher2013recursive}. 
\paragraph{Topic Classification} AGNews (news domain) and Yahoo Answers (Yahoo; web domain)~\citep{zhang2015character}.

\paragraph{Details} We experiment on the modified test set used in~\citet{shi2022nearest,min2022nonparametric}.\footnote{They made a random subset of 3,000 samples for datasets with larger original test sets.}
We report the scores averaged over four templates~\citep{gao2021making, min2022noisy}. See Table~\ref{table:verbalizer_setup} for the list of verbalizers used.

\subsection{Multiple Choice QA Datasets}
We evaluate on RACE-M (R-M), RACE-H (R-H)~\citep{lai2017race}, ARC-E, ARC-C~\citep{clark2018think}, Open Book Question Answering (OBQA)~\citep{mihaylov2018can}, and CommonsenseQA (CoQA)~\citep{talmor2018commonsenseqa}. 
For CoQA, we report the validation results as the official test set is not publicly available.
We use a single template from \citet{holtzman2021surface}. See Table~\ref{tab:verbalizer_multichoice} for details on templates and verbalizers.

\subsection{Baseline Models}
Standalone decoder language models \textbf{(LM)} select the label whose word sequence has with the highest probability. We use \textit{gpt-2-large}~\citep{radford2019language} and variants of \textit{gpt-3}~\citep{brown2020language}.
\textbf{PMI}~\cite{holtzman2021surface} calibrates the decoder models with a domain-conditioned premise. 
We also implement encoder-based cloze-style zero-shot classification task~\cite{gao2021making} \textbf{(LM-BFF)} with \textit{roberta-large}~\citep{liu2019roberta} as the encoder.

For retrieval-augmented baselines, we use \textbf{kNN-LM}~\cite{khandelwal2019generalization} and \textbf{kNN-prompt}~\cite{shi2022nearest}. These models adjust the output token probabilities with external resources. 


\citet{yin2019benchmarking} regards label candidates as hypotheses and predicts the label with the highest entailment logit score \textbf{(Entailment)}. 
For implementation, we finetune pre-trained \textit{roberta-large}~\citep{liu2019roberta} on NLI (SNLI + MNLI) datasets~\citep{bowman2015snli,williams2018mnli}.\footnote{We use the \href{https://github.com/UKPLab/sentence-transformers/tree/master/examples/training/cross-encoder}{script} provided by sentence-transformers~\citep{reimers2019sentence}.}





\subsection{Implementation Details}

\paragraph{Sentence Encoder}
\label{paragraph:sentence_encoder}
For the main experiments, We use two off-the-shelf sentence encoder models: \textit{sup-simcse-roberta-large} (SimCSE)~\citep{gao2021simcse} trained on NLI, and \textit{sentence-t5-large} (ST5)~\citep{ni2022sentence} trained on CommunityQA and NLI. Among the ST5 variations, we use an encoder-only mean version.
We also experiment with larger versions of ST5 (\textit{sentence-t5-xl} (ST5-XL) and \textit{sentence-t5-xxl} (ST5-XXL)) to see the effect of model size.

\paragraph{External Corpus}
\label{paragraph:external_corpus}
We collect the external corpus used in~\citet{shi2022nearest} which consists of Wiki103, IMDB, subsets of CC-News, and Amazon Review.

\paragraph{Retrieval}
We use the same sentence encoders as Section~\ref{paragraph:sentence_encoder} for building the embedding index, and use FAISS~\citep{johnson2019billion} for the top-k computation and retrieval.
For each original label prompt, we use 5 synonymous prompts and retrieve 5 sentences per query (so that K=25) in closed-set classification. See Table~\ref{table:retrieved_samples_sentiment} and Table~\ref{table:retrieved_samples_agn} for a list of retrieved examples.
For multiple choice tasks, we use 25 retrieved sentences for a single query. We do not use synonymous label prompts as making small changes in the wordings may deviate from the originally intended semantics, thus making the option less of a viable answer.
We observe no further gain beyond 25 retrieved samples.

\section{Results}\label{sec:results}


Our method shows solid performance on both task types under zero-shot setting.

\subsection{Closed-set Classification}
\label{res:closed_set}
Table~\ref{tab:main_close_set} shows the results for closed-set classification tasks.

\paragraph{Baselines}
On average, KNN-prompt shows the strongest results among all baselines. This indicates the importance of extracting and exploiting relevant information from external knowledge sources in zero-shot inference.
Among the baselines that do not make use of external resources, LM-BFF outperforms the standalone decoder models with or without PMI calibration, presumably due to utilizing bi-directional information with an encoder.
\paragraph{\ours}
Our method shows strong performance when coupled with a well-built sentence encoder. With the ST5 backbones, we outperform all the baselines despite fewer parameters.\footnote{We attribute ST5's strong zero-shot classification abilities to its architecture rather than the pretraining dataset composition. See appendix~\ref{appendix:nli_st5} for details.} Changing the sentence encoder to SimCSE results in a lower performance, but still comparable to LM-BFF. 

\ours~serves as a cost-effective method of leveraging retrieval from external sources. Although KNN-Prompt~\citep{shi2022nearest} depends on the same external corpus as ours, \ours~outperforms KNN-Prompt not only in terms of task accuracy\footnote{\ours$_{\text{ST5}}$ (335M) reaches a higher performance than KNN-Prompt which has twice as more parameters (774M)).}, but also in terms of computational efficiency. Specifically, whereas KNN-Prompt requires token-level encoding of the leftward context and the next token, we employ a much lighter instance-level encoding. Such difference in the encoding strategy allows~\ours~to operate on merely 2M items while KNN-Prompt needs to store 284M items, taking external corpus mentioned in Section \ref{paragraph:external_corpus}. 

\begin{table}[t]
\small
\renewcommand{\arraystretch}{1.2}
\begin{tabular}{lc}
\toprule
\textbf{Retrieval Strategy} & \textbf{Avg} \\
\hline
\textit{ST5} \\
no retrieval & 77.4 \\
retrieval \textbackslash w single query & 78.4 \\
retrieval \textbackslash w multiple synonymous queries & 78.9   \\
\hline
\textit{SimCSE} \\
no retrieval &  73.6 \\
retrieval \textbackslash w single query &  74.7 \\
retrieval \textbackslash w multiple synonymous queries &  75.6 \\
\bottomrule
\end{tabular}
\caption{Retrieval ablation results.}
\label{tab:ret_ablation}
\vspace{-4mm}
\end{table}

\begin{table*}[t]
\centering
\small
\resizebox{1.0\textwidth}{!}{
\renewcommand{\arraystretch}{1.2}
\begin{tabular}{lrccccccc}
\toprule
 \textbf{Method} & \textbf{\# Params} & \textbf{R-M} & \textbf{R-H} & \textbf{ARC-E}& \textbf{ARC-C} & \textbf{OBQA} & \textbf{CoQA} & \textbf{Avg} \\
\hline
\textbf{Baselines (Language Modeling)} \\
GPT-2  & 774M & 39.3  & 31.8  & 52.7    & 23.1    & 19.4   & 33.3    & 33.3    \\
\addlinespace[2.5pt]
\multirow{3}{*}{GPT-3$^\dagger$} & 6.7B & 43.3 & 34.8 & 58.2 & 26.8 & 22.4 & 40.0 & 37.6     \\
  & 13B & 49.6 & 38.2 & 66.2 & 32.1 & 28.2 & 48.8 & 43.9\\
  & 175B    & \underline{55.7}  & 42.4  & \underline{73.5}    & 40.2   & 33.2   & 61.0    & 51.0   \\
\hdashline
\addlinespace[2.5pt]
GPT-2 + PMI~\citep{holtzman2021surface}   & 774M & 43.9  & 38.3  & 47.0    & 31.6    & 43.2   & 44.5    & 41.4    \\
\addlinespace[2.5pt]
\multirow{3}{*}{GPT-3 + PMI$^\dagger$~\citep{holtzman2021surface}} & 6.7B & 48.5 & 39.8 & 51.5 & 33.0 & 48.0 & 50.3 & 45.2 \\
   & 13B & \textbf{51.3} & 42.1 & 57.7 & 38.5 & 50.4 & 58.5 & 49.8 \\
   & 175B   & \underline{55.7}  & \underline{43.7}  & 63.3    & \underline{45.5}   & \underline{58.0}   & \underline{66.7}    & \underline{55.5} \\

\hline
\textbf{Baselines (NLI format)} \\
Entailment~\citep{yin2019benchmarking} & 355M & 39.1  & 29.9  & 45.9    & 31.1    & 42.8   & 35.1    & 37.3    \\
\hline
\textbf{Ours (Sentence Encoder)} \\
\ours$_\text{SimCSE}$ & 355M & 39.4  & 35.1  & 48.3  & 26.5  & 38.2   & 47.3    & 39.1    \\
\ours$_\text{ST5}$ & 335M & 40.5  & 38.3  & 56.2    & 28.7    & 44.4   & 55.5    & 43.9 \\
\ours$_\text{ST5-XL}$ & 1.24B & 43.7  & 40.3  & 63.8    & 37.0    & 48.4   & 59.0    & 48.7 \\
\ours$_\text{ST5-XXL}$ & 4.8B & 45.9  & \textbf{42.3}  & \textbf{69.3}    & \textbf{44.5}    & \textbf{52.8}   & \textbf{63.9}    & \textbf{53.1} \\

\bottomrule
\end{tabular}
}
\caption{Results for multiple choice tasks under zero-shot setting.
The \textbf{Boldface} indicates the best performances in each column except the models with 175B parameters.
The \underline{Underline} shows the best scores in each column regardless of model size.
We reproduce GPT-2 and GPT-2 + PMI with the code provided by \citet{holtzman2021surface}. The symbol $\dagger$ indicates the performance reported by \citet{holtzman2021surface}.
}
\label{tab:main_open_set}
\end{table*}

\paragraph{Ablation on Label Prompt Augmentation}
\label{subsection:retrieval_strategy}
We evaluate the effects of using additional label prompts on top of the original prompt. From Table~\ref{tab:ret_ablation}, we observe that using supplementary label information results in solid performance increase.  Our full augmentation setting where multiple synonymous prompts are used as retrieval queries (``retrieval \textbackslash w multiple synonymous queries'') shows a solid performance increase over the ablative baselines of a narrower form of retrieval (``retrieval \textbackslash w single query'') or none at all (``no retrieval'').
This is especially true for topic classification tasks; in particular, removing the retrieval component in AGNews decreases the accuracy by 6.2 points ($76.6 \rightarrow 70.4$)~\footnote{For performance gap on other datasets, see Appendix~\ref{appendix:ret_ablation}}.

\subsection{Multiple Choice QA}
\label{res:open_set}
Table~\ref{tab:main_open_set} shows the results for multiple choice QA tasks.


\paragraph{Baselines} 
In sharp contrast to the results in Section~\ref{res:closed_set}, the decoder approaches show strong performance with dramatic gains from upscaling. The highest scores come from a giant decoder model (175B GPT-3+PMI), presumably due to massive memorization of specific knowledge.  

\paragraph{\ours} 
Still, our method based on sentence encoders still achieves better performance than the baselines with an equivalent (or larger) number of parameters.
\ours$_{\text{ST5}}$ (with 335M parameters) outperforms GPT-2+PMI (which has the best performance with $\le$ 774M parameters) by 2.5 points in average. Our best model \ours$_{\text{ST5-XXL}}$ (with 4.8B parameters) outperforms much larger GPT-3+PMI baselines (with 6.7B and 13B parameters) by 7.9 and 3.3 points in average; even on par with a 175B GPT-3 despite having 36.5x fewer parameters. 

\section{Analysis}\label{sec:analysis}\label{section:analysis}

\subsection{Verbalizer Sensitivity Test}
\label{subsection:sensitivity}

\begin{figure*}[t]
\resizebox{\textwidth}{!}{
\includegraphics{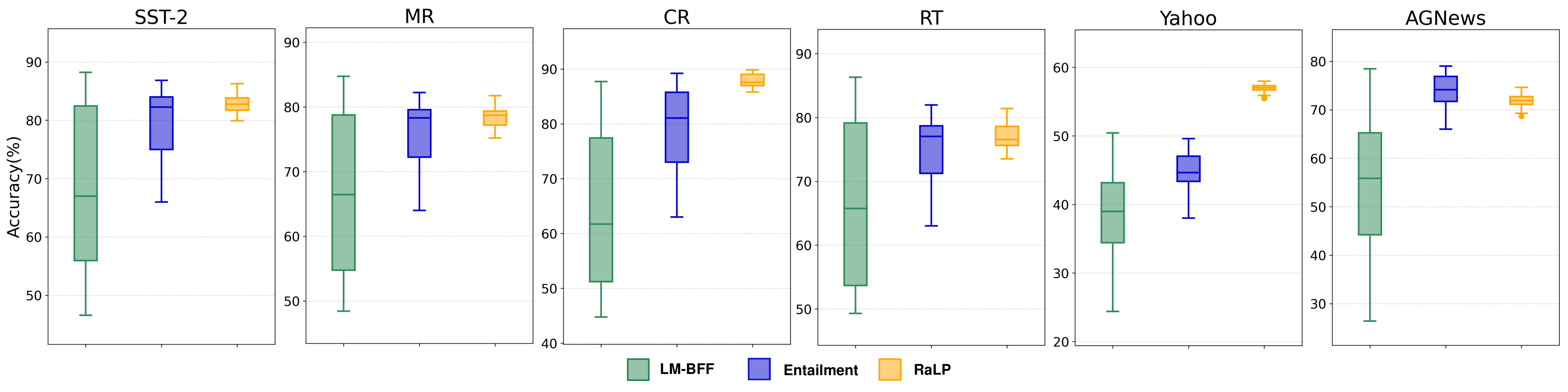}
}
\caption{ Results with varying verbalizers~(\ref{subsection:sensitivity}) in closed-set classification datasets.
The middle line in each plot marks the mean score across all templates. 
Our approach has smaller variance to template variations than \citet{gao2021making,yin2019benchmarking}.
}
\label{fig:violin_plot_sensitivity}
\end{figure*}

Existing works based on prompting are vulnerable to verbalizer changes, with worst-case performances often down to chance-level~\citep{lu2022fantastically}.
We suggest that using sentence encoders may be a remedy, as they are trained to distribute semantically similar samples nearby during contrastive pre-training and thus could be less sensitive to surface form variations. In this section, we empirically verify that our method based on sentence encoders is \emph{reliably} strong.

\paragraph{Settings}
We measure the variance in performance across a range of paraphrased label templates while keeping the label words intact:
say, changing \textit{``It was great.''} for sentiment analysis to \textit{``It's a great thing.''} or \textit{``That's great.''}.
For paraphrasing, we leverage templates from existing works~\citep{gao2021making, min2022noisy} and augmentation techniques~\citep{ma2019nlpaug}
including back-translation~\citep{sennrich2016backtransalation} and contextual word embedding-based augmentation~\citep{kobayashi2018contextualaug}.
Among the generated candidates, we manually filter out the ones with semantic distortions (\emph{e.g.} added negations).\footnote{See Table~\ref{table:verbalizer_varation_setup} for a full list.}
We use \emph{roberta-large} for the baselines~\citep{yin2019benchmarking, gao2021making}. For fair comparison, we share the same backbone with the baselines and use \textit{sup-simcse-roberta-large} (SimCSE)~\citep{gao2021simcse} as our sentence encoder.


\paragraph{Results}
Figure~\ref{fig:violin_plot_sensitivity} shows the accuracy distributions over verbalizer variations.
As previously reported~\citep{jiang2021know}, cloze-style inference~\citep{gao2021making} is overly sensitive to subtle contextual modifications.
The Entailment~\citep{yin2019benchmarking} approach is less volatile, but it still suffers from sporadic performance drops with certain templates as shown in long lower whiskers in the plots.
On the other hand, \ours~has more stable performance distribution under template variations, and this holds for all datasets. 

\begin{figure*}[t]
\centerline{\includegraphics[width=\textwidth]{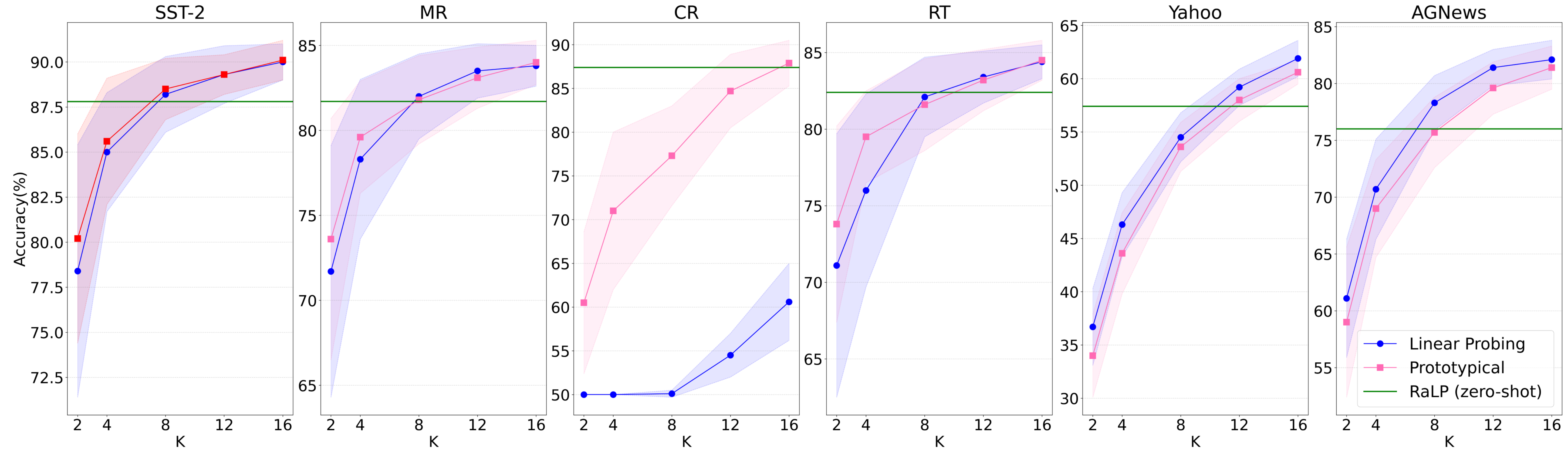}}
\caption{
Performances of two few-shot approaches (mean and standard deviation across 50 runs) compared to \ours.
The x-axis represents the number of training examples per class.
All use embeddings from \textit{sentence-t5-large}.
}
\label{fig:fewshot_plot}
\end{figure*}

\subsection{Comparison with Few-shot Setup}

If we allow \emph{some} supervision (\emph{i.e.} few-shot), there exist alternative strategies that can be used in combination with sentence encoder representations: linear probing and prototypical networks~\citep{snell2017prototypical,dopierre2021neural}. Therefore, we figure out whether our zero-shot approach is still attractive as compared to the two few-shot options. 
\paragraph{Baselines} 
Linear probing requires training the classification head (\textit{i.e.} logistic classification). 
The prototypical network treats the average embeddings of support examples as  \textit{class prototypes} and measures the distance of instance from these prototypes. 
For both methods, we randomly choose K=$\{2,4,8,12,16\}$ samples per class from training examples with 50 different seeds and report the averaged score.

\paragraph{Results}
Figure~\ref{fig:fewshot_plot} illustrates that both few-shot learning baselines are volatile on sample selection variations.
Specifically, linear probing exhibits more variability when given a small number of labeled samples.
This may be because linear probing requires training parameters from scratch.
Compared to few-shot training methods, \ours~shows robust performance when the supervision is extremely limited (i.e., K $\in [2, 4, 8]$). 



\section{Related Work}\label{sec:related_work}

\paragraph{Prompting}
\citet{brown2020language} reformulate downstream tasks as language modeling with prompting, where the approach performs remarkably in few or zero-shot setup and even shows competitive performance to supervised methods~\citep{schick2021exploiting, jiang2021know}.
Subsequent works validate that even small language models can make few or zero-shot inferences with prompting~\citep{gao2021making,schick2021just,le2021many,logan2022cutting}.
However, the performance of prompting methods varies greatly depending on which prompt is used~\citep{perez2021true, lu2022fantastically}.
To address this issue, a line of works explore calibrating the predictions of language models for the few-shot inference~\citep{holtzman2021surface,zhao2021calibrate}.
Unlike previous prompting methods relying on language modeling (\textit{i.e.} token prediction), our framework, which separately encodes label prompts and text input, is particularly robust to surface form variations.

\paragraph{Zero-shot Classification}
Prompting methods for zero-shot inference have been actively studied.
These works use the likelihood of verbalizer given input text~\citep{brown2020language,holtzman2021surface,zhao2021calibrate}, the conditional probability of the input text given verbalizer~\citep{min2022noisy}, and masked language model objective~\citep{gao2021making} for solving NLU tasks.
To facilitate zero-shot inference, \citet{shi2022nearest} incorporates information from an external corpus using a fuzzy verbalizer to adjust the decoding logits.
NPM~\citep{min2022nonparametric}, on the other hand, removes the softmax function and only utilizes an external corpus to retrieve the answer.

Another line of research formulates the classification as a natural language inference (NLI) task.
\citet{yin2019benchmarking} solves the classification task by comparing the entailment logit scores where they concatenate the text input and label prompt as text pairs.
\citet{gera2022zero} further improves the framework by finetuning the model using self-training on unlabeled train data.
SimPTC~\citep{fei2022beyond}, concurrent to our work, propose solving classification using sentence encoder and clustering method.
Their method is similar to our approach without a retrieval component but differs from our work in that (1) it requires a chunk of unlabeled test dataset for clustering, (2) it does not use a retrieval component, (3) it cannot be expanded to other tasks such as multiple choice QA.

\paragraph{Sentence Encoder}
Sentence encoders project the sentences to embedding space which can be applied to various language understanding tasks.~\citep{reimers2019sentence, du2021self, gao2021simcse, ni2022sentence}.
Generally, sentence encoders are trained by contrastive representation learning, where the encoder learns to distribute semantically similar samples nearby in embedding space.
Hence, representations obtained by sentence encoders are well-suited features for classification in limited supervision setup~\citep{tunstall2022efficient}
However, existing methods based on sentence encoders train an additional classification head to solve downstream tasks, hindering them from using fine-grained representation and being unsuitable for zero-shot inference.
In contrast, our approach utilizes label prompts for the zero-shot classification which does not incur additional parameters and fully exploits fine-grained representations.

\section{Conclusion}\label{sec:conclusion}This paper introduces \ours, a metric-based classification framework where the input under inference is compared to candidate class anchors in metric space.
\ours~uses the label prompts as class anchors in an embedding space of a sentence encoder pretrained with a contrastive learning objective.
To compensate for underspecified label names, we retrieve the pseudo-label prompts from external knowledge sources and refine the prediction scores.
We verify that \ours~is a powerful zero-shot classification strategy through an extensive evaluation on six closed-set text classification tasks and six open-set multiple-choice QA tasks.
Further performance boost by retrieval demonstrates that class anchor selection is important for zero-shot classification and can be readily and stably achieved by label prompt augmentation.

\section{Limitations}\label{sec:limitation}
Though \ours~demonstrates its strong performance in zero-shot classification tasks, the approach has several limitations.
First, it is non-trivial to apply our method to non-classification tasks such as generation.
Hence, future works can explore adapting the sentence encoders or their training methods in developing language models which have been explored in the context of the retrieval-augmented language models~\citep{zhong2022training}.
Moreover, while our work demonstrates its robust performance compared to some of the few-shot learning methods such as linear probing or prototypical network, we did not cover finetuning methods for our approach.
Future works can explore tuning the parameters of the sentence encoders when given limited labeled data~\cite {tunstall2022efficient}.


\bibliography{anthology,custom}
\bibliographystyle{acl_natbib}

\clearpage
\appendix

\section*{Appendix}\label{sec:appendix}


\section{Analyzing Performance Gap between SimCSE and ST5}
\label{appendix:nli_st5}


\begin{table*}[t]
\centering
\small
\renewcommand{\arraystretch}{1.2}
\begin{tabular}{lccccccccc}
\toprule
{Method} & {Dataset} & {SST-2} & {MR} & {CR} &  {RT} & {Yahoo} & {AGNews} & {Avg}\\
\hline
\ours$_\text{ST5}$   & CommunityQA + NLI & \textbf{87.8}& \textbf{81.7} & 87.4 & \textbf{82.4} & 57.4 & 76.6    &\textbf{78.9} \\
\ours$_\text{ST5}$   & NLI & 85.7 & 80.2 & \textbf{87.8} &80.0 & \textbf{58.8} & \textbf{76.8} & 78.2  \\
\ours$_\text{SimCSE}$   & NLI & 82.3 & 78.0 & 87.1 & 76.7 & 57.0 & 72.6 & 75.6 \\
\bottomrule
\end{tabular}
\caption{Results for ablation study of pre-training dataset. We additionally trained the ST5 model using only NLI dataset.}
\label{tab:appendix_nli_st5}
\end{table*}


In Section~\ref{res:closed_set}, we evaluate the performance of various pre-trained sentence encoders using \ours.
Our results show that \ours$_{\text{SimCSE}}$~(\textit{sup-simcse-roberta-large}) has a lower performance than \ours$_{\text{ST5}}$~(\textit{sentence-t5-large} although they have the same parameter size.
We identify two main differences between SimCSE and ST5: i) the pre-trained model utilized as the initial checkpoint, and ii) the utilization of solely the NLI datasets in SimCSE as opposed to a combination of NLI and CommunityQA in ST5.
Since the CommunityQA dataset is not publicly available, analyzing the reasons for this performance gap is important for future research on sentence encoders for zero-shot text classification.
To do so, we conduct experiments by comparing the two encoder models.

For a fair comparison, we train the ST5 model using only the NLI datasets.
Following the original setting in \citet{ni2022sentence}, we employ T5-large ~\citep{raffel2020t5} as the initial pre-trained model and evaluate its performance through zero-shot text classification datasets.
As shown in Table~\ref{tab:appendix_nli_st5}, the performance of the \ours$_\text{ST5}$ trained with the combination of CommunityQA and NLI achieved comparable performance with \ours$_\text{ST5}$ model trained using only the NLI dataset.
However, among models trained using only NLI, \ours$_\text{ST5}$ outperforms \ours$_\text{SimCSE}$ by 2.6 points on average.
This suggests that the performance gap observed in Table~\ref{tab:main_close_set} is primarily influenced by the T5 architecture rather than the CommunityQA dataset.
We assume that the objective of reconstructing the consecutive span of corrupted tokens from one unique mask token~(\textit{i.e.} span masking) used in the pre-training of T5 has enhanced the model's general-purpose knowledge more than the objective of BERT-style~(\textit{i.e.} token masking).


\begin{table*}[t]
\centering
\small
\renewcommand{\arraystretch}{1.1}
\begin{tabular}{lrcccccccc}
\toprule
\textbf{Method} & \textbf{\# Params} & \textbf{SST-2} & \textbf{MR} & \textbf{CR} &  \textbf{RT} & \textbf{Yahoo} & \textbf{AGNews} & \textbf{Avg}\\
\hline
\textbf{Ours (Sentence Encoder)} \\
\ours$_\text{SimCSE}$ & 355M & 82.3 & 78.0        & 87.1        & 76.7        & 57.0 & 72.6    &75.6 \\
- no retrieval  & 355M & 80.8 & 76.5 & 87.5 & 75.7 & 53.4 & 67.6 & 73.6 \\
\ours$_\text{ST5}$               & 335M & 87.8 & 81.7        & 87.4   & 82.4        & 57.4 & 76.6   & 78.9 \\
- no retrieval  & 355M & 87.7 &	81.7  &	87.3	& 81.9	& 55.3 & 	70.4 & 77.4 \\



\bottomrule
\end{tabular}
\caption{Retrieval ablation full results.
}
\label{tab:ret_ablation_full}
\end{table*}

\begin{table*}[t]
\centering
\resizebox{0.8\textwidth}{!}{
\renewcommand{\arraystretch}{1.2}
\begin{tabular}{ll}
\toprule
\textbf{Dataset}           & \textbf{Verbalizers}\\
\hline

SST-2, MR, CR, RT & A \texttt{MASK} one.; It was \texttt{MASK}.; All in all \texttt{MASK}.; A \texttt{MASK} piece.                       \\
                  & (\texttt{MASK} = \{great, terrible\})                                                                         \\
\hline
AGNews            & Topic: \texttt{MASK}.; Subject: \texttt{MASK}.; This is about \texttt{MASK}.; It is about \texttt{MASK}.               \\
                  & (\texttt{MASK} = \{World, Sports, Business, Technology\}) \\
\hline
Yahoo             & (Same as above) (\texttt{MASK} = \{Company, Educational Institution, Artist, Athlete,  \\ &
Office Holder, Mean of Transportation, Building, Natural Place, Village, Animal,\\
                  & Plant, Album, Film, Written Work\}) \\
\bottomrule
\end{tabular}
}
\caption{The details of verbalizer setting in text classification task. We follow the verbalizers from \citet{gao2021making,min2022noisy}}
\label{table:verbalizer_setup}
\end{table*}

\begin{table*}[t]
	\centering
        \resizebox{0.8\textwidth}{!}{
        \renewcommand{\arraystretch}{1.2}
	\begin{tabularx}{\textwidth}{l X X}
    \toprule
		\textbf{Dataset} & \textbf{Premise} $x$ & \textbf{Hypothesis} $y$ \\
        \hline




        \textbf{RACE} & There is not enough oil in the world now. As time goes by, it becomes less and less, so what are we going to do when it runs out[...].] \textcolor{blue}{question:} According to the passage, which of the following statements is true\textcolor{blue}{?}  & \textcolor{blue}{answer:} There is more petroleum than we can use now. \\ \hline

		\textbf{ARC}  & What carries oxygen throughout the body? & \textcolor{blue}{the answer is:} red blood cells. \\ \hline
		
        \textbf{OBQA}  & Which of these would let the most heat travel through? & \textcolor{blue}{the answer is:} a steel spoon in a cafeteria. \\ \hline

		\textbf{CoQA}  & Where can I stand on a river to see water falling without getting wet? & \textcolor{blue}{the answer is:} bridge. \\ 
    \bottomrule
    \end{tabularx}
    }
\caption{
The details of verbalizer setting in multiple choice datasets. 
The hypothesis candidates are directly given as label prompt in each task. 
The texts with \textcolor{blue}{blue color} denote templates that we prepend following \citet{holtzman2021surface}}.
\label{tab:verbalizer_multichoice}
\end{table*}

\begin{table*}[b]
\centering
\resizebox{0.8\textwidth}{!}{
\renewcommand{\arraystretch}{1.2}
\begin{tabular}{l l}
\toprule
\textbf{Dataset}           & \textbf{Verbalizers}\\
\hline

SST-2, MR, CR, RT &         
        It was \texttt{MASK}.;
        A \texttt{MASK} piece.;
        A \texttt{MASK} one.; \\ &
        All in all \texttt{MASK}.;  
        It was absolutely \texttt{MASK}, really.; \\ &
        It was really so \texttt{MASK}.; 
        It was more than just \texttt{MASK}.; \\ & 
        And that was absolutely \texttt{MASK}, too.;
        It was still pretty \texttt{MASK}.; \\ &
        It was all \texttt{MASK}.; 
        It was literally \texttt{MASK}.;
        It's a \texttt{MASK} thing.; \\ &
         What a \texttt{MASK} performance.; 
        It is an \texttt{MASK} piece at the best of times.; \\ &
        One \texttt{MASK} piece.; 
        A \texttt{MASK} work.; 
        A \texttt{MASK} play.; \\ &
        This is a \texttt{MASK} one.;
        It is utterly \texttt{MASK}.; 
        A really \texttt{MASK} one.; \\ &
        That's \texttt{MASK}.;
        That's all \texttt{MASK}.;
        All told this is a truly \texttt{MASK} thing.; \\ &
        A \texttt{MASK} overall.;
        All together \texttt{MASK}.
                       \\
                  & (\texttt{MASK} = \{great, terrible\})                                                                         \\
\hline
AGNews            & Topic: \texttt{MASK}.;
        Subject: \texttt{MASK}.;
        This is about \texttt{MASK}.;  \\ &
        It is about \texttt{MASK}.;  
        It's about the \texttt{MASK}.; \\ &
        It's about \texttt{MASK}.;
        It's all about the \texttt{MASK}.; \\ &
        It's just about the \texttt{MASK}.;
        It's the whole lot with the \texttt{MASK}.; \\ &
        Here, we are talking about the \texttt{MASK}.;
        This is the theme of the \texttt{MASK}.; \\ &
        It is related to the \texttt{MASK}.;
        It is about what it means for the \texttt{MASK}.; \\ &
        This involves the \texttt{MASK}.; 
        Theme: \texttt{MASK}.; \\ &
        keyword: \texttt{MASK}.;
        On a related topic: the \texttt{MASK}.; \\ &
        It is for the \texttt{MASK}.;
        The subject: the \texttt{MASK}.; \\ &
        Main topic: \texttt{MASK}.;
        Content: \texttt{MASK}.;\\ &
        Theme is the \texttt{MASK}.;
        Issue: \texttt{MASK}.; \\ &
        Executive Summary: \texttt{MASK}.;
        Material: \texttt{MASK}.               \\ &
                  (\texttt{MASK} = World, Sports, Business, Technology)                                                               \\ 
\hline
Yahoo             & (Same as above) (\texttt{MASK} = \{Company, Educational Institution, Artist, Athlete,  \\ &
Office Holder, Mean of Transportation, Building, Natural Place, Village, Animal,\\
                  & Plant, Album, Film, Written Work\}) \\
\bottomrule
\end{tabular}
}
\caption{We use all listed verbalizers in verbalizer variation test in Section~\ref{subsection:sensitivity}. We evaluate on total 25 verbalizer with or without punctuation.}
\label{table:verbalizer_varation_setup}
\end{table*}
\begin{table*}[b]
\centering
\resizebox{1.0\textwidth}{!}{
\renewcommand{\arraystretch}{1.2}
\begin{tabular}{l l}
\toprule
\textbf{Original Label Prompt}           & \textbf{Retrieval Augmented Label Prompts}\\
\hline

It was great. 
& great, love it - great, love it - great, love it - great, love it really did love it and it was also great ;\\
& It was really good! I reallly liked it! As does my father and mother and brother and sister-in-law. It was epic. ;\\
& i loved it. it could not have been better. justin kirby ;\\
& it was awesome, maiden is at there best, is was totally awesomeness, nobody comes close to maiden, it's a totally fulfillment ;\\
& it was awesome!!!!! of course it was awesome. its Harry Potter for lord's sake!!!! \\

\hdashline
It was good.
& A <unk> . " A is good " , or " A was good " . ;\\ 
& great, love it - great, love it - great, love it - great, love it really did love it and it was also great ;\\ 
& It was ok. Wish I could give this a great review, sorry. But, it was still ok, if you're interested. ;\\ 
& It was alright having not seen it for many years i enjoyed watching it and am glad for the purchase ;\\ 
& It was really good! I reallly liked it! As does my father and mother and brother and sister-in-law. It was epic. \\

\hdashline
It was famous.
& = = = In popular memory = = = ;\\ 
&  = = Famous people who witnessed it = = ;\\ 
&  = = = National and international fame = = = ;\\ 
& This is a real classic, with a number of actors and actresses that were to become very famous later on. ;\\ 
& This classic was very innovative for it's time and is a classic.The Princess created quite a stir at the time, and a cult following. \\

\hline
It was terrible.
& terrible, terrible.It wasm't worth the money or the time to watch it. we turned it off in the middle of it ;\\ 
& It was horrible. One of the worst books ever. Thank youBelly Up ;\\ 
& This book was terrible and was horrific when it came to the description. ;\\ 
& It was awful. I went into it with great hopes expecting to enjoy it. I wanted to like it, I wanted to laugh. I spent more time rolling \\ & my eyes and checking the counter to see how much time was left till it was over. I was so disappointed. ;\\
& Just poor quality and I just didn't like it.Really couldn't pay attention because it was terrible, awful movie I wouldn't recommend. \\

\hdashline
It was horrible.
& It was horrible. One of the worst books ever. Thank youBelly Up ;\\ 
& This book was terrible and was horrific when it came to the description. ;\\ 
& It was sick, depressing, annoying, and not to mention rotten! so that's what I have to say on the subject! ;\\ 
& I hated it and thought it was really bad. It was twisted and is really scary to litte children ;\\ 
& truth be told, this is a horrible story of what really happened - so very awful...it was impossible to read \\
\hdashline
It was awful.
& It was horrible. One of the worst books ever. Thank youBelly Up ;\\ 
& It was awful. I went into it with great hopes expecting to enjoy it. I wanted to like it, I wanted to laugh. I spent more time rolling \\ & my eyes and checking the counter to see how much time was left till it was over. I was so disappointed. ;\\ 
& This book was terrible and was horrific when it came to the description. ;\\ 
& It was sick, depressing, annoying, and not to mention rotten! so that's what I have to say on the subject! ;\\ 
& The movie was truly a wretched experience. The six letters to describe it are S-U-C-K-E-D! \\
\bottomrule
\end{tabular}
}
\caption{Original label prompt in sentiment analysis task~(SST-2, MR, CR, and RT) and retrieval augmented label prompt in Section~\ref{res:closed_set}.}
\label{table:retrieved_samples_sentiment}
\end{table*}
\begin{table*}[b]
\centering
\resizebox{0.8\textwidth}{!}{
\renewcommand{\arraystretch}{1.2}
\begin{tabular}{l l}
\toprule
\textbf{Original Label Prompt}           & \textbf{Retrieval Augmented Label Prompts}\\
\hline
Topic: world.
& = = = = Worldwide = = = =  ;  
  = = = = Informing the world = = = =  ;\\ 
&  = = = On the Wisdom of this World = = =  ; 
  = State of the World ( song ) =  ;\\ 
&  = = = = HELLO , WORLD = = = =  
\\
\hdashline
Topic: international.
& = = = = International = = = =  ; 
  = = = = International affairs = = = =  ;\\ 
&  = = = = International services = = = =  ; 
  = = = = International relations = = = =  ;\\ 
&  = = = International ( InLine ) = = = 
\\

\hdashline
Topic: global.
& = = = = Worldwide = = = =  ; 
  = = = Global environmental impact = = =  ;\\ 
&  = = = From the global community = = =  ; 
  = = = Around the globe = = =  ;\\ 
&  = = = = Terrestrial globes = = = =
\\

\hline
Topic: sports.
& = = = = Sports = = = =  ; 
  = = = = Athletic history = = = =  ;\\ 
&  = = = Effects on sports = = =  ; 
  = = = World sport context = = =  ;\\ 
&  = = = = Sport = = = =
\\

\hdashline
Topic: entertainment.
& = = = = Entertainment = = = =  ; 
  = = = Entertainment and culture = = =  ;\\ 
&  Entertainment is a form of activity that holds the attention and interest of an audience , \\ & or gives  pleasure and delight  . It can be [...] and even for a global audience .  ;\\ 
&  = = = Sports and entertainment = = =  ; 
  = = = Media and entertainment = = =
\\

\hdashline
Topic: recreation.
& = = = = Recreations = = = =  ; 
  = = = = Recreation = = = =  ;\\ 
&  = = = Sports and recreation = = =  ; 
  = = = Sport and recreation = = =  ;\\ 
&  = = = = Tourism and recreation = = = = 
\\

\hline
Topic: business.
& = = = = Business = = = =  ;
  = = = = On labor and business = = = =  ;\\ 
&  = = = Lines of business = = =  ; 
  = = = DE 1 Business = = =  ;\\ 
&  = = = Operations and business = = =
\\

\hdashline
Topic: economics.
&  = = = = Economic lecture = = = =  ; 
  = = = = Economics = = = =  ;\\ 
&  = = = Views on economics = = =  ; 
  = = = = Economic = = = =  ;\\ 
&  = = = Philosophy of economics = = =  
\\

\hdashline
Topic: financial.
& = = = = Finances = = = =  ; 
  = = = Financial and loan = = =  ;\\ 
&  = = = = Fiscal issues = = = =  ; 
  = = = = Financial performance = = = =  ;\\ 
&  = = = Economy and finance = = =  
\\

\hline
Topic: technology.
&  = = = Influence on technology = = =  ; 
  = = = Views on the technology = = =  ;\\ 
&  = = = = Technical aspects = = = =  ; 
  = = = Effects of technology = = =  ;\\ 
&  = = = Technology and science = = = 
\\

\hdashline
Topic: science.
& Science – news on science @-@ related topics ( e.g. cool technology , space telescope \\ & observations , interesting medical research ) .  ;\\ 
&  = = = Interactions with the scientific community = = =  ; 
  = = = = Scientific uses = = = =  ;\\ 
&  = = = History of science = = =  ; 
  = = = Science and scientism = = = 
\\

\hdashline
Topic: mathmatics.
&  = = = Attention towards mathematics = = =  ; 
  = = = = Mathematics = = = =  ;\\ 
&  = = = An Introduction to Mathematics = = =  ;
  = = = Mathematics as an art = = =  ;\\ 
&  = = = Philosophy of mathematics = = =
\\
\bottomrule
\end{tabular}
}
\caption{Original label prompt for AGNews and retrieval augmented label prompt in Section~\ref{res:closed_set}.}
\label{table:retrieved_samples_agn}
\end{table*}

\section{Evaluation Protocol}
We report the accuracy for all datasets. 
To optimize hyperparameters, we exploit the validation dataset for each task. In case of CoQA, we sub-sample 3,000 instances from the training dataset, regarding them as in-house validation examples. 

\section{Full Results on Retrieval Ablation Study}
\label{appendix:ret_ablation}

We report the full results of the retrieval ablation study in Table~\ref{tab:ret_ablation_full}.

\section{Implementation Details}
In all experiments, we evaluate models on a single A100 with 80GB, and A6000 GPU with 48GB of memory. 
We implement all models with PyTorch using sentence-transformers library from UKPLab\footnote{https://github.com/UKPLab/sentence-transformers}.
We choose the best hyperparameter of top-\textit{K} in $\{5, 10, 25, 50, 100\}$ for \ours~ based on validation split.

\end{document}